\documentclass{article}
\usepackage{spconf,amsmath,graphicx,hyperref}
\usepackage{multirow}
\usepackage{booktabs}
\usepackage{xcolor}
\definecolor{lblue}{rgb}{0.95, 0.94, 0.99}
\title{RFOP: Rethinking Fusion and Orthogonal Projection for \\ Face-Voice Association}

\name{Abdul Hannan$^{1}$, Furqan Malik$\textsuperscript{2}$, Hina Jabbar$^3$, Syed Suleman Sadiq$\textsuperscript{*}$, Mubashir Noman$^4$}
\address{$^{1}$University of Trento, Italy \quad $^{2}$Saxion University of Applied Sciences, Netherlands \\ $^{3}$University of Education, Lahore, Pakistan \quad $^4$MBZUAI, UAE}

\begin{document}
%\ninept
%
\maketitle
\begin{abstract}
Face-voice association in multilingual environment challenge 2026 aims to investigate the face-voice association task in multilingual scenario.
The challenge introduces English-German face-voice pairs to be utilized in the evaluation phase. 
To this end, we revisit the fusion and orthogonal projection for face-voice association by effectively focusing on the relevant semantic information within the two modalities. Our method performs favorably on the English-German data split and ranked \textbf{3rd} in the FAME 2026 challenge by achieving the EER of $33.1$. Our code and models are available at \url{https://github.com/techmn/rfop}.
\end{abstract}

\section{Introduction}
\label{sec:intro}
% \vspace{-3mm}
%The task of face voice association is introduced by Nagrani et al. \cite{nagrani2018seeing} which intends to identify whether the given pair of face and voice belong to the same or different identity.
The face voice association task verify whether the given face and voice pair belong to the same or different identity \cite{nagrani2018seeing,nawaz2019deep}.
Recently, face-voice association in multilingual environment (FAME) challenge \cite{saeed2024synopsis,moscati2025face} is pushing it to the next level by exploring the association of a face in multilingual scenario. The FAME challenge calls into the following question: \textit{Can the model trained on seen-heard pairs perform equally well on unseen-unheard identities of the same and different language}. Apart from the face-voice pairs for English and Urdu languages, the current FAME challenge includes the face-voice pairs from English and German languages, posing significant challenges for the cross-modal verification task. 

Existing methods \cite{nagrani2018learnable, nawaz2021cross, horiguchi2018face,saeed2023single, shah2023speaker, chen2023local} utilize domain specific encoders to extract feature representations from face and voice pairs. Later, these methods minimize the feature distance between the same identities by means of triplet or contrastive loss functions. However, these methods need optimal selection of margin hyperparameter and become more computationally expensive as the size of the data increases. 
Consequently, the baseline method (FOP) \cite{saeed2022fusion} demonstrated the effectiveness of combining the orthogonal loss with Cross Entropy loss for cross-modal verification. 
More recently, PAEFF \cite{hannan25_interspeech} illustrated that the well-constructed fusion module and the alignment of multimodal representations before fusion improve the face voice association performance. These methods rely on enhancing the fusion approaches and the losses to bring the same identities closer, since the unimodal feature extractors are frozen and the preprocessing of samples is not properly applied to handle the noise. Given that the FAME'26 provides the pre-extracted features for efficient training, we adapted the FOP as our baseline method. We observed that the recent PAEFF aligns the multimodal representations in hyperbolic domain for precise alignment, however, such alignment works better when the samples are noise free. Moreover, we empirically observed that the gated fusion \cite{saeed2022fusion} is not effective when combined with feature alignment. 
Therefore, it was desired to rethink the design of multimodal feature fusion while effectively aligning the face and voice representations. 

%###########################################
\begin{figure*}[!t]
    \centering
    \includegraphics[width=0.80\linewidth]{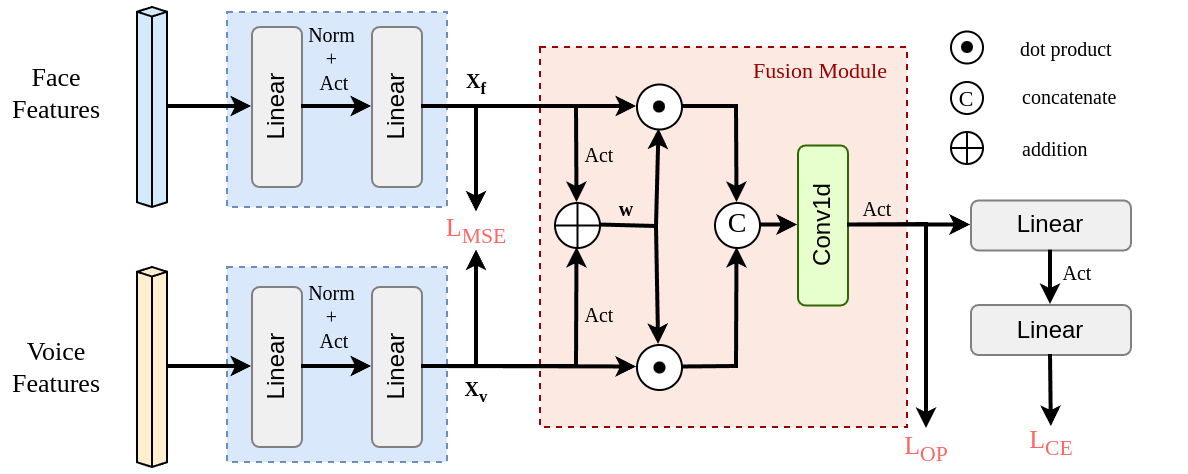}
    \caption{Overall face-voice association network. Face and voice features are extracted from corresponding unimodal encoders and projected by means of linear projection layers obtaining embeddings $X_f$ and $X_v$, which are combined by fusion module to obtain fused embeddings. 
    %Finally, these embeddings are passed to the header to obtain class probabilities. 
    Finally, the network is optimized using combination of three losses given as: $L_{total}=\alpha_1 L_{MSE} + \alpha_2 L_{OP} + \alpha_3 L_{CE}$. }
    \label{fig:overall_framework}
    % \vspace{-5mm}
\end{figure*}

\section{Method}
\label{sec:method}
%\vspace{-3mm}
We illustrate the architecture of the proposed face voice association method in Fig.~\ref{fig:overall_framework}. The method aims to address the shortcomings of the baseline FOP by focusing on the relevant semantic information in both modalities. The proposed framework takes the face and voice representations extracted from the corresponding encoders as input. Each modality representations are projected to the latent embedding space having same dimensions by using two linear layers to obtain features $X_f$ and $X_v$, respectively. Given that the multimodal representations are in the latent space, mean squared error (MSE) loss is utilized for global alignment of the embeddings. The latent embeddings $X_f$ and $X_v$ are then passed to the fusion module to combine them. 

In fusion module, we first obtain the attention weights $w$ by applying the non-linear activation function on latent embeddings followed by addition. Subsequently, the attention weights are multiplied with embeddings $X_f$ and $X_v$ to highlight the relevant significant information. We then concatenate the features and feed them to the convolution layer for feature mixing. 

Finally, the output of the fusion module is projected by a linear header to obtain the class probabilities. Similar to the baseline, we utilize orthogonal projection loss on the fused embeddings to minimize the distance between same identities while maximizing the separability between the different class identities. Moreover, Cross Entropy loss is utilized to measure the performance of the model during training. The final objective is to optimize the following loss function:

\begin{equation}
    L_{total}=\alpha_1 L_{MSE} + \alpha_2 L_{OP} + \alpha_3 L_{CE}
\end{equation}

\noindent where $\alpha_1, \alpha_2$ and $\alpha_3$ are hyperparameters.
%\vspace{-3mm}

\section{Experiments}
%\vspace{-3mm}
\textbf{Implementation Details: }
%\subsection{Implementation Details}
The framework is implemented using pytorch library on a single RTX $6000$ GPU. We utilized AdamW optimizer to optimize the model having weight decay of $0.2$. We set initial learning rate to $0.01$ and decayed its value by using cosine annealing schedule. During training, we use the batch size of $64$ and trained the model for $100$ epochs in two phases. In first phase, we train the model for $50$ epochs using hyperparameters as discussed above. Afterwards, we select the best performing model on the small validation set and initialize the model with these weights. We further train the model for $50$ epochs using learning rate of $0.0001$ and select the best performing model for reporting.

In our experiments, we empirically select the values of $\alpha_1, \alpha_2$ and $\alpha_3$ as 0.02, 0.78 and 0.2, respectively. We observed that the large value of $\alpha_2$ provides better separation between the dissimilar identities, however, further increasing its value after a certain point degrades the cross-modal verification results. 

\noindent\textbf{Metrics: } According to the challenge criteria, we utilize the \textit{equal error rate} (EER) to measure the performance of the model. 

% \subsection{Results}
\noindent \textbf{Results: }
We present the best performance of our model in Table~\ref{tab:fame26_v3}. We observe that the performance of the model is better when it is trained and tested on the heard language, validating that the model is able to recognize the semantics of the same language. In contrast, the performance of the model is considerably degraded when it is trained on the English language while tested on the German language. This reveals that there is a large semantic difference between the two languages which hinders the ability of the model to easily adapt to the new language. However, we notice that the performance difference is less when the model is trained on German language while tested on English language. We conjecture that this low performance difference is due to the presence of English language samples within the German face-voice pairs which enables the model to learn some of the English language semantics from the data. 
%##########################################
%\begin{table}
%\caption{ Cross-modal verification on FAME'26 dataset-v1 containing English and Urdu pairs. }
%\centering
%\resizebox{1.0\linewidth}{!}{
%\begin{tabular}{l|c|cc|c}
%\hline
%\multirow{2}{*}{Method}  & \multirow{2}{*}{Configuration} & \multicolumn{3}{c}{EER $\downarrow$}  \\
%\cline{3-5}

%&  & English Test & Urdu Test & Overall Score \\
%\hline
%\multirow{2}{*}{FOP~\cite{saeed2022fusion}} & English Train & 29.3 & 37.9 &  \multirow{2}{*}{33.4} \\
%& Urdu Train & 40.4	& 25.8 &  \\
%\hline
%\multirow{2}{*}{RFOP (ours)}  & English Train & 33.8 & 29.0 & \multirow{2}{*}{31.4}  \\
%& Urdu Train & 41.7 & 21.0 &  \\
%\hline
%\end{tabular}
%}
%\label{tab:fame26_v1}
%\end{table}
%##########################################

%##########################################
\begin{table}
\caption{Cross-modal verification on FAME'26 dataset-V3 containing English and German languages. }
\vspace{0.2cm}
\centering
\resizebox{1.0\linewidth}{!}{
\begin{tabular}{l|c|cc|c}
\hline
\multirow{2}{*}{Method}  & \multirow{2}{*}{Configuration} & \multicolumn{3}{c}{EER $\downarrow$}  \\
\cline{3-5}

&  & English Test & German Test & Overall Score \\
\hline
\multirow{2}{*}{FOP~\cite{saeed2022fusion}} & English Train & 35.3 & 48.0 &  \multirow{2}{*}{41.5} \\
& German Train & 45.1 & 37.9 &  \\
\hline
\multirow{2}{*}{RFOP (ours)}  & English Train & 25.4 & 41.1 & \multirow{2}{*}{33.1}  \\
& German Train & 34.7 & 31.2 &  \\
\hline
\end{tabular}
}
\label{tab:fame26_v3}
\end{table}
%##########################################

\bibliographystyle{IEEEbib}
\small
\bibliography{refs}

@inproceedings{nagrani2018learnable,
  title={Learnable pins: Cross-modal embeddings for person identity},
  author={Nagrani, Arsha and Albanie, Samuel and Zisserman, Andrew},
  booktitle={ECCV},
  pages={71--88},
  year={2018}
}

@inproceedings{nawaz2019deep,
  title={Deep latent space learning for cross-modal mapping of audio and visual signals},
  author={Nawaz, Shah and Janjua, Muhammad Kamran and Gallo, Ignazio and Mahmood, Arif and Calefati, Alessandro},
  booktitle={2019 Digital Image Computing: Techniques and Applications (DICTA)},
  pages={1--7},
  year={2019},
  organization={IEEE}
}

@inproceedings{nagrani2018seeing,
  title={Seeing voices and hearing faces: Cross-modal biometric matching},
  author={Nagrani, Arsha and Albanie, Samuel and Zisserman, Andrew},
  booktitle={CVPR},
  pages={8427--8436},
  year={2018}
}

@inproceedings{shah2023speaker,
  title={Speaker recognition in realistic scenario using multimodal data},
  author={Shah, Saqlain Hussain and Saeed, Muhammad Saad and Nawaz, Shah and Yousaf, Muhammad Haroon},
  booktitle={2023 3rd International Conference on Artificial Intelligence (ICAI)},
  pages={209--213},
  year={2023},
  organization={IEEE}
}

@inproceedings{saeed2023single,
  title={Single-branch network for multimodal training},
  author={Saeed, Muhammad Saad and Nawaz, Shah and Khan, Muhammad Haris and Zaheer, Muhammad Zaigham and Nandakumar, Karthik and Yousaf, Muhammad Haroon and Mahmood, Arif},
  booktitle={ICASSP},
  pages={1--5},
  year={2023},
  organization={IEEE}
}

@article{moscati2025face,
  title={Face-voice Association in Multilingual Environments (FAME) 2026 Challenge Evaluation Plan},
  author={Moscati, Marta and Abdullah, Ahmed and Saeed, Muhammad Saad and Nawaz, Shah and Das, Rohan Kumar and Zaheer, Muhammad Zaigham and Mir, Junaid and Yousaf, Muhammad Haroon and Malik, Khalid and Schedl, Markus},
  journal={arXiv preprint arXiv:2508.04592},
  year={2025}
}

@inproceedings{saeed2024synopsis,
  title={A synopsis of fame 2024 challenge: Associating faces with voices in multilingual environments},
  author={Saeed, Muhammad Saad and Nawaz, Shah and Moscati, Marta and Das, Rohan Kumar and Tahir, Muhammad Salman and Zaheer, Muhammad Zaigham and Liaqat, Muhammad Irzam and Khan, Muhammad Haris and Nandakumar, Karthik and Yousaf, Muhammad Haroon and others},
  booktitle={ACM Multimedia},
  pages={11333--11334},
  year={2024}
}

@inproceedings{saeed2022fusion,
  title={Fusion and orthogonal projection for improved face-voice association},
  author={Saeed, Muhammad Saad and Khan, Muhammad Haris and Nawaz, Shah and Yousaf, Muhammad Haroon and Del Bue, Alessio},
  booktitle={ICASSP},
  pages={7057--7061},
  year={2022},
  organization={IEEE}
}

@inproceedings{horiguchi2018face,
  title={Face-voice matching using cross-modal embeddings},
  author={Horiguchi, Shota and Kanda, Naoyuki and Nagamatsu, Kenji},
  booktitle={ACM Multimedia},
  pages={1011--1019},
  year={2018}
}

@inproceedings{nawaz2021cross,
  title={Cross-modal speaker verification and recognition: A multilingual perspective},
  author={Nawaz, Shah and Saeed, Muhammad Saad and Morerio, Pietro and Mahmood, Arif and Gallo, Ignazio and Yousaf, Muhammad Haroon and Del Bue, Alessio},
  booktitle={CVPR},
  pages={1682--1691},
  year={2021}
}

@inproceedings{chen2023local,
  title={Local-Global Contrast for Learning Voice-Face Representations},
  author={Chen, Guangyu and Zhang, Deyuan and Liu, Tao and Du, Xiaoyong},
  booktitle={ICIP},
  pages={51--55},
  year={2023},
  organization={IEEE}
}

@inproceedings{hannan25_interspeech,
  title     = {{PAEFF: Precise Alignment and Enhanced Gated Feature Fusion for Face-Voice Association}},
  author    = {Abdul Hannan and Muhammad Arslan Manzoor and Shah Nawaz and Muhammad Irzam Liaqat and Markus Schedl and Mubashir Noman},
  year      = {2025},
  booktitle = {Interspeech },
  pages     = {2710--2714},
  doi       = {10.21437/Interspeech.2025-268},
  issn      = {2958-1796}
}

\end{document}